\documentclass[conference]{IEEEtran}
\IEEEoverridecommandlockouts
\usepackage{cite}
\usepackage{amsmath,amssymb,amsfonts}
\usepackage{algorithmic}
\usepackage{graphicx}
\usepackage{textcomp}
\usepackage{xcolor}
\def\BibTeX{{\rm B\kern-.05em{\sc i\kern-.025em b}\kern-.08em
    T\kern-.1667em\lower.7ex\hbox{E}\kern-.125emX}}

\usepackage{hyperref}

\usepackage{booktabs}
\usepackage{multirow}
\usepackage{threeparttable}
\DeclareMathOperator*{\argmin}{arg\,min}

\begin{document}

\title{A Comparative Study of Deep Learning Architectures for Multi-Horizon Behavioural Forecasting for Mobile Health}

\author{
	\IEEEauthorblockN{
	    Pavlos Nicolaou\textsuperscript{a}, Kleanthis Malialis\textsuperscript{a}, Artemis Kontou\textsuperscript{a}, and Panayiotis Kolios\textsuperscript{a, b}
	}
	\IEEEauthorblockA{
		\textsuperscript{a} \textit{KIOS Research and Innovation Center of Excellence}\\
		\textsuperscript{b} \textit{Department of Computer Science}\\
		\textit{University of Cyprus},
		Nicosia, Cyprus
        \thanks{Email:\texttt{\{nicolaou.pavlos, malialis.kleanthis, kontou.artemis, kolios.panayiotis\}@ucy.ac.cy}}
        \thanks{ORCID: 0000-0001-8520-0105 (PN), 0000-0003-3432-7434 (KM), 0000-0003-0480-0796 (AK), 0000-0003-3981-993X (PK)}
        \thanks{This work was supported by the European Union’s Horizon 2020 research and innovation programme under grant agreement No 739551 (KIOS CoE TEAMING) and from the Republic of Cyprus through the Deputy Ministry of Research, Innovation and Digital Policy. It was also supported by the CIPHIS (Cyprus Innovative Public Health ICT System) project of the \mbox{NextGenerationEU} programme under the Republic of Cyprus Recovery and Resilience Plan under grant agreement C1.1l2.}
        }
}

\maketitle

\begin{abstract}

Wearable devices and smartphones generate rich behavioural time series that can support proactive health interventions, yet systematic comparisons of modern forecasting architectures for these data are lacking. In particular, it remains unclear how models generalise across populations, how different architectures respond to participant-level fine-tuning and how forecasting accuracy degrades across multi-day horizons. We benchmark six deep learning architectures, two zero-shot Foundation Models (FM) and statistical baselines on three public datasets encompassing over 800 participants, reporting per-feature metrics for step counts, screen time and sleep duration across 1-8 day horizons. We further conduct a per-feature personalisation study across all six architectures and assess FM transferability across dataset sizes and temporal granularities. Our key findings are: (i) no single architecture dominates, PatchTST leads among trained models while the three runners-up (TCN, MLP, Transformer) show no meaningful performance difference; (ii) the FM TimesFM matches or exceeds trained models zero-shot, especially in low-data regimes and (iii) participant-level fine-tuning reduces per-feature RMSE by 16-60\%, with sleep benefiting most and step counts least. These results provide practical guidance on architecture selection, FM applicability and personalisation strategies for mobile health forecasting. To the best of our knowledge, this is the first study to jointly evaluate modern deep learning, FMs and personalisation for multi-horizon behavioural forecasting from wearables.

\end{abstract}


\section{Introduction}
\subsection{Motivation}
Wearable devices and smartphones have become ubiquitous tools for continuous health monitoring, collecting streams of physiological and behavioural data, including physical activity, sleep patterns, screen usage and communication patterns \cite{sabry2022machine, chen2025passive}. These passively collected data streams offer unprecedented opportunities for predictive health applications that can support proactive interventions \cite{baig2024leveraging}. Machine learning, particularly deep learning, has demonstrated remarkable capabilities in extracting meaningful patterns from such complex sensor data, enabling applications ranging from depression detection to activity recognition \cite{chen2022deep_har, jacobson2025depression}.

Multi-horizon forecasting, predicting health-related features at multiple future time steps, represents a particularly valuable paradigm for personalised health applications~\cite{lim2021temporal, fan2019multihorizon}. Unlike single-step predictions, multi-horizon forecasts enable anticipatory interventions, allowing individuals and healthcare providers to identify potential behavioural changes before they occur or in early stages. This capability shifts health support from reactive monitoring to anticipatory care, a predicted multi-day decline in physical activity could flag an emerging depressive episode before mood symptoms come to surface~\cite{wang2018tracking}, while forecasted sleep disruption might prompt medication timing adjustments for a patient managing a chronic condition. From the individual's perspective, multi-horizon forecasts provide actionable foresight, not a blank statement, for example "you slept poorly last night" but a clear understanding of the sleep is predicted to decline over the next three days, enabling proactive steps such as scheduling an intervention or requesting a medical expert to check in. Such timely, forecast-driven nudges leverage behavioural momentum and are more effective than reactive corrections issued after the behaviour has already occurred. At the population level, accurate multi-day forecasts can inform resource allocation in digital health programmes and enable adaptive intervention delivery \cite{baig2024leveraging}.

\subsection{Challenges}

Despite the growing body of work in both time series forecasting and wearable health sensing, several critical gaps remain. First, while Transformer-based models and state-space models such as Mamba have demonstrated strong performance across diverse forecasting domains \cite{vaswani2017attention, gu2024mamba}. Nevertheless, there is a lack of systematic comparison of these modern architectures specifically for health-related behavioural features from wearables. Prior studies in mobile health (mHealth) sensing have predominantly focused on classification tasks, for example emotion or depression detection, rather than continuous forecasting \cite{xu2022globem_neurips, xu2022globem_imwut, cornet2023wearable}

Second, the challenge of cross-population generalisation remains largely unsolved. Prior work has shown that models trained on one population often fail to generalise to new data collected from different people, with performance degradation rates that generally exceed 18\% \cite{xu2022globem_neurips, xu2022globem_imwut}. This finding highlights the importance of personalisation strategies, yet systematic evaluation of how different forecasting architectures respond to personalisation is lacking.

Third, most existing work evaluates models using single-step or short-horizon predictions \cite{lim2021time}, while practical health applications require understanding how prediction accuracy degrades across multiple horizons (for example, 1 to 8 days ahead). This multi-horizon perspective is essential for determining the utility window of different forecasting approaches.

\subsection{Contributions}
\begin{itemize}
    \item We benchmark six deep learning architectures, two zero-shot FMs and statistical baselines on three public datasets, reporting per-feature metrics for step counts, screen time and sleep duration across 1-8 day horizons.
    \item We present per-feature, per-architecture personalisation study for behavioural forecasting from wearables, quantifying how fine-tuning gains vary by feature regularity and how they reshape architecture rankings.
    \item We provide evaluation of time-series FMs on mHealth data, comparing zero-shot performance against trained architectures across dataset sizes and temporal granularities.
\end{itemize}

The rest of the paper is organised as follows. Section~\ref{sec:related-work} reviews related work. Section~\ref{sec:methodology} describes the methodology, including datasets, architectures and personalisation. Section~\ref{sec:experimental-setup} details the experimental setup. Section~\ref{sec:results} presents results across architectures, horizons and personalisation strategies. Section~\ref{sec:discussion} discusses findings and practical implications, and Section~\ref{sec:conclusion} concludes the paper.

\section{Related Work}
\label{sec:related-work}

\subsection{Deep Learning for Time Series Forecasting in Mobile Health}
\label{sec:deep-learning}
Deep learning has progressively supplanted classical statistical methods for time-series forecasting, offering the capacity to learn complex nonlinear temporal dependencies directly from data~\cite{lim2021time, torres2021deep}. Recurrent Neural Networks, particularly Long Short-Term Memory (LSTM) networks~\cite{hochreiter1997lstm}, were among the earliest deep models adopted for sequential health data, with probabilistic extensions such as DeepAR demonstrating their utility for autoregressive forecasting~\cite{salinas2020deepar}. Temporal Convolutional Networks (TCNs) introduced a parallel-processing alternative through dilated causal convolutions, matching or exceeding LSTM accuracy on several benchmarks while training faster~\cite{bai2018empirical, lara2020tcn}. In clinical settings, TCNs have proved effective for mortality and length-of-stay prediction~\cite{bednarski2022temporal}.

The Transformer~\cite{vaswani2017attention} brought global self-attention to time-series modelling. PatchTST improved efficiency by segmenting the input into subseries-level patches and processing each channel independently \cite{nie2023patchtst}. State-space models have emerged as linear-complexity alternatives: Mamba introduces input-dependent selection to the structured state-space framework~\cite{gu2024mamba} and adaptations such as S-Mamba and MambaTS have demonstrated competitive forecasting accuracy at a fraction of Transformer cost~\cite{wang2024smamba, cai2025mambats}.

Despite these advances, the mobile-health literature remains dominated by shallow classifiers (Random Forests, SVMs and single-layer Neural networks), applied primarily to detection rather than forecasting tasks \cite{chen2025passive, cornet2023wearable, fawaz2019deep}. A systematic comparison of modern deep architectures for continuous behavioural forecasting from wearable and smartphone data is, to our knowledge, still lacking.

\subsection{Foundation Models for Time Series in Mobile Health}

A recent paradigm shift has seen the emergence of large pre-trained FMs for time-series analysis. TimesFM \cite{das2024timesfm} is a decoder-only model pre-trained on over 100 billion real-world time points, enabling zero-shot univariate forecasting across arbitrary horizons without task-specific training.

Reverso \cite{fu2026reverso} takes an efficiency-first approach, employing small hybrid architectures that interleave long convolution and linear recurrent (DeltaNet) layers. Despite being orders of magnitude smaller than Transformer-based alternatives, Reverso pushes the performance-efficiency Pareto frontier through tailored data augmentation and inference strategies.

These models are primarily used in a \emph{zero-shot} setting, where the pre-trained model is applied directly without task-specific training. Some, such as TimesFM, also support \emph{fine-tuning} selected layers on target data. While both modes perform well on standard benchmarks such as energy and traffic forecasting, it remains unclear whether they generalise to health domains, where data exhibit high inter-individual variability, irregular sampling, and small cohort sizes.

\subsection{Personalisation Strategies in Mobile Health}

Individual variability in behavioural patterns is a defining characteristic of wearable health data~\cite{sabry2022machine, li2024comparison}. Population-level (global) models trained on pooled data often fail to capture patterns from each individual participant and the GLOBEM benchmark has shown that cross-population generalisation degrades accuracy by over 18\% on average~\cite{xu2022globem_neurips, xu2022globem_imwut}. While GLOBEM provides an important benchmark for mobile sensing and cross-population generalisation, its primary focus is on predictive modelling for mental health-related outcomes and generalisation across populations. In contrast, the present study focuses specifically on continuous multi-horizon behavioural forecasting of wearable-derived features, compares modern forecasting architectures and time-series FMs and evaluates participant-level fine-tuning across multiple behavioural features and forecast horizons

A simple personalisation strategy is \emph{personalised fine-tuning}: a global model is first trained on major part of the dataset and then updated on each user's records~\cite{jeong2025personalized}. This has yielded accuracy improvements exceeding 25\% for emotion recognition from wearables \cite{li2024comparison} and substantial gains for depression prediction \cite{xu2021collaborative}. However, fine-tuning requires sufficient per-user data, with too few samples, it could be overfitting to noise. 

\section{Methodology}
\label{sec:methodology}
\subsection{Problem Formulation}

Let
\begin{equation}
\mathbf{X}_{1:T} = (\mathbf{x}_1, \mathbf{x}_2, \ldots, \mathbf{x}_T), \quad \mathbf{x}_t \in \mathbb{R}^D
\end{equation}
be a multivariate time series with $D$ features recorded over $T$ time steps. Given an input window of length $L$, define the multivariate input matrix:
\begin{equation}
\mathbf{X}_{t-L+1:t} \in \mathbb{R}^{L \times D}.
\end{equation}
The objective is to predict a multivariate future sequence over forecast horizon $H$:
\begin{equation}
\mathbf{Y}_{t+1:t+H} = (\mathbf{x}_{t+1}, \ldots, \mathbf{x}_{t+H}) \in \mathbb{R}^{H \times D}.
\end{equation}
The forecasting problem reduces to estimating a parametric mapping
\begin{equation}
f_\theta : \mathbb{R}^{L \times D} \to \mathbb{R}^{H \times D},
\end{equation}
such that $\hat{\mathbf{Y}}_{t+1:t+H} = f_\theta(\mathbf{X}_{t-L+1:t})$.

Define the pooled training set $\mathcal{D} = \bigcup_{u \in \mathcal{U}_{\text{train}}} \mathcal{D}_u$.
We consider three modelling paradigms, Global, Foundation and Personalised.

\textbf{Global.} A single model $f_{\theta^*}$ is trained on the
pooled dataset by minimising the Huber loss $\mathcal{L}$:
\[
\theta^* = \argmin_\theta \sum_{(\mathbf{X}, \mathbf{Y}) \in \mathcal{D}}
\mathcal{L}\bigl(f_\theta(\mathbf{X}),\, \mathbf{Y}\bigr),
\]
and evaluated on held-out users via a participant-disjoint split
($\mathcal{U}_{\text{train}} \cap \mathcal{U}_{\text{eval}} = \emptyset$).

\textbf{Foundation.} A pre-trained model $f_{\theta_0}$ is applied zero-shot, i.e.\ $\theta = \theta_0$, without any task-specific optimisation on the target data.

\textbf{Personalised.} Starting from the global optimum $\theta^*$, we fine-tune on each user's data:
\[
\theta_u^* = \argmin_\theta \sum_{(\mathbf{X}, \mathbf{Y}) \in \mathcal{D}_u}
\mathcal{L}\bigl(f_\theta(\mathbf{X}),\, \mathbf{Y}\bigr),
\quad \theta^{(0)} = \theta^*.
\]

Let $\mathcal{U} = \{u_1, \ldots, u_N\}$ denote a population of $N$ users.
For each user $u \in \mathcal{U}$, let
$\mathcal{D}_u = \{(\mathbf{X}^{(i)}_u, \mathbf{Y}^{(i)}_u)\}_{i=1}^{n_u}$
denote their set of sliding-window input--output pairs, where
$\mathbf{X}^{(i)}_u \in \mathbb{R}^{L \times D}$ is a lookback window of $L$ days and
$\mathbf{Y}^{(i)}_u \in \mathbb{R}^{H \times D}$ is the corresponding forecast horizon.

\subsection{Personalisation Strategy}

We adopt a two-stage transfer learning procedure \cite{jeong2025personalized}: (1) pre-train a global model to obtain $\theta^*$ on $\mathcal{D}$, then (2) fine-tune on $\mathcal{D}_u$ for each evaluation user $u \in \mathcal{U}_{\text{eval}}$ to obtain $\theta_u^*$. All six trained architectures are personalised under identical conditions. We compare the global ($\theta^*$) and personalised ($\theta_u^*$) variants, reporting per-participant metrics across all forecast horizons.

\label{sec:experimental-setup}
\section{Experimental Setup}

\subsection{Datasets and Features}
We evaluate on three publicly available datasets:
\begin{itemize}
    \item \textbf{GLOBEM}~\cite{xu2022globem_neurips, xu2022globem_imwut, goldberger2000physiobank}: Multi-year mobile sensing data accessed through PhysioNet. The dataset contains daily behavioural features extracted from smartphones and Fitbit wearables across four years (2018-2021). We split participants into 382 training, 95 validation and 200 evaluation users.
    \item \textbf{CAPTURE-24}~\cite{chan2024capture}: Wrist-worn accelerometer data from 151 participants sampled at 100\,Hz, aggregated to hourly activity summaries.
    \item \textbf{Continuous Wearables Dataset}~\cite{baigutanova2025continuous}: Daily step count and sleep duration from 49 participants wearing consumer wrist devices over several weeks.
\end{itemize}

\subsection{Compared Methods}

We benchmark six deep learning architectures, two FMs alongside three statistical baselines. All trained models share a common sliding-window input and a shared sequence decoder, so performance differences reflect the encoder. Table \ref{tab:complexity} summarises model sizes.

\smallskip\noindent\textbf{Trained architectures:}
\begin{itemize}
    \item \textbf{MLP}: Feedforward network, a simple non-temporal baseline.
    \item \textbf{LSTM} \cite{hochreiter1997lstm}: Gated recurrent model for sequential dependencies.
    \item \textbf{TCN}~\cite{bai2018empirical}: Dilated causal convolutions with parallel processing.
    \item \textbf{Transformer}~\cite{vaswani2017attention}: Global self-attention over the full window.
    \item \textbf{Mamba}~\cite{gu2024mamba}: State-space model with linear complexity.
    \item \textbf{PatchTST}~\cite{nie2023patchtst}: Patch-level Transformer with channel independence.
\end{itemize}

\smallskip\noindent\textbf{Foundation models (zero-shot):}
\begin{itemize}
    \item \textbf{TimesFM}~\cite{das2024timesfm}: Decoder-only model pre-trained on 100B+ time points.
    \item \textbf{Reverso}~\cite{fu2026reverso}: Compact hybrid of long convolutions and linear recurrence.
\end{itemize}

\smallskip\noindent\textbf{Statistical baselines:}
\begin{itemize}
    \item \textbf{Naive (Last)}: Repeats the last observed value for all horizon steps.
    \item \textbf{Moving Average}: Rolling mean over a 4-observation window.
    \item \textbf{ARIMA}: Auto-regressive integrated moving average with automatic order selection.
\end{itemize}

\begin{table}[ht]
\centering
\caption{Model complexity. Parameter counts for GLOBEM ($D$=3, $w$=8, $h$=8). Avg.\ epochs to early stopping (patience 100).}
\label{tab:complexity}
\small
\begin{tabular}{lrr}
\toprule
\textbf{Model} & \textbf{Params} & \textbf{Avg.\ epochs} \\
\midrule
TCN & 98K & 200 \\
PatchTST & 152K & 319 \\
MLP & 218K & 188 \\
LSTM & 368K & 147 \\
Reverso$^\dagger$ & 550K & --- \\
Transformer & 1.7M & 140 \\
Mamba & 2.0M & 139 \\
TimesFM$^\dagger$ & 200M & --- \\
\bottomrule
\multicolumn{3}{l}{\footnotesize $^\dagger$Foundation model (zero-shot, no training).}
\end{tabular}
\end{table}

\subsection{Evaluation Method}

All models are trained and evaluated under a single, standardised protocol to ensure fair comparison. Each dataset is split by participant into non-overlapping train, validation, and evaluation subsets: GLOBEM uses 382/95/200, CAPTURE-24 uses 105/23/23 and Wearables uses 10 held-out evaluation users. Evaluation participants are never seen during
training.

All trained models minimise the Huber loss ($\delta{=}1.0$), optimised with AdamW (learning rate $10^{-3}$, weight decay $10^{-4}$) for up to 200 epochs with early stopping and cosine annealing. Evaluation proceeds differently under each paradigm:

\begin{itemize}
    \item \textbf{Global.} A single model trained on pooled data from all training users is applied directly to each evaluation user's sliding windows.
    \item \textbf{Foundation.} The pre-trained model is applied zero-shot with a fixed context window. No gradient updates are performed on the target data.
    \item \textbf{Personalised.} Starting from the global checkpoint, the model is fine-tuned on each evaluation user's data with a reduced learning rate ($10^{-4}$) for up to 200 epochs (early stopping, patience 100).
\end{itemize}

All metrics (Section~\ref{sec:metrics}) are computed per participant and reported as mean$\pm$std across evaluation users.

\subsection{Performance Metrics}
\label{sec:metrics}
We report four complementary metrics, each computed per participant and summarised as mean$\pm$std across the evaluation cohort.

\textbf{RMSE} (Root Mean Squared Error):
\begin{equation}
\text{RMSE} = \sqrt{\frac{1}{N}\sum_{i=1}^{N}(y_i - \hat{y}_i)^{2}}
\end{equation}
where $y_i$ and $\hat{y}_i$ are the actual and predicted values for the $i$-th sample and $N$ is the total number of predictions. Primary metric due to its sensitivity to large errors, which are particularly consequential in health applications.

\textbf{MAE} (Mean Absolute Error):
\begin{equation}
\text{MAE} = \frac{1}{N}\sum_{i=1}^{N}|y_i - \hat{y}_i|
\end{equation}
Outlier-robust complement to RMSE.

\textbf{sMAPE} (symmetric Mean Absolute Percentage Error):
\begin{equation}
\text{sMAPE} = \frac{100}{N} \sum_{i=1}^{N} \frac{|y_i - \hat{y}_i|}{(|y_i| + |\hat{y}_i|)/2}
\end{equation}
Scale-independent accuracy measure bounded between 0 and 200\%. Unlike MAPE, sMAPE remains defined when actual values are zero, a common occurrence for step counts on sedentary days.

\textbf{Skill Score}:
\begin{equation}
\text{Skill} = 1 - \frac{\text{RMSE}_{\text{model}}}{\text{RMSE}_{\text{NaiveLast}}}
\end{equation}
Positive values indicate improvement over NaiveLast and negative values indicate worse performance.

\section{Results}
\label{sec:results}
\subsection{Per-Feature Architecture Comparison}

\begin{table*}[t]
\centering
\caption{Global and foundation model results on GLOBEM at $h$=8 ($w$=8 for trained models; foundation models use 64-day context). Values are mean{\tiny$\pm$}std of per-participant metrics ($N$=200 for trained and foundation models).}
\label{tab:per_feature_h8_global}
\scriptsize
\begin{tabular*}{\textwidth}{@{\extracolsep{\fill}} l rrrr rrrr rrrr}
\toprule
& \multicolumn{4}{c}{\textbf{Steps (daily count)}} & \multicolumn{4}{c}{\textbf{Screen (min/day)}} & \multicolumn{4}{c}{\textbf{Sleep (min/day)}} \\
\cmidrule(lr){2-5} \cmidrule(lr){6-9} \cmidrule(lr){10-13}
\textbf{Model} & RMSE & MAE & sMAPE & Skill & RMSE & MAE & sMAPE & Skill & RMSE & MAE & sMAPE & Skill \\
\midrule
\multicolumn{13}{l}{\textit{Foundation (zero-shot)}} \\
TimesFM & 1363{\tiny$\pm$580} & --- & --- & 0.28{\tiny$\pm$0.46} & 39{\tiny$\pm$19} & --- & --- & 0.23{\tiny$\pm$0.38} & 42{\tiny$\pm$53} & --- & --- & $-$0.20{\tiny$\pm$1.51} \\
Reverso & 2073{\tiny$\pm$642} & --- & --- & $-$0.09{\tiny$\pm$0.50} & 45{\tiny$\pm$19} & --- & --- & 0.12{\tiny$\pm$0.37} & 41{\tiny$\pm$50} & --- & --- & $-$0.16{\tiny$\pm$1.44} \\
\midrule
\multicolumn{13}{l}{\textit{Global (trained)}} \\
PatchTST & \textbf{631}{\tiny$\pm$245} & \textbf{386}{\tiny$\pm$134} & \textbf{26.4}{\tiny$\pm$11.7} & \textbf{0.67}{\tiny$\pm$0.14} & \textbf{21}{\tiny$\pm$8} & \textbf{14}{\tiny$\pm$7} & \textbf{18.4}{\tiny$\pm$8.4} & \textbf{0.57}{\tiny$\pm$0.20} & \textbf{34}{\tiny$\pm$34} & 21{\tiny$\pm$18} & 6.8{\tiny$\pm$8.3} & \textbf{0.01}{\tiny$\pm$0.11} \\
MLP & 836{\tiny$\pm$381} & 557{\tiny$\pm$252} & 37.0{\tiny$\pm$14.5} & 0.55{\tiny$\pm$0.18} & 26{\tiny$\pm$14} & 19{\tiny$\pm$11} & 24.0{\tiny$\pm$9.2} & 0.45{\tiny$\pm$0.27} & 39{\tiny$\pm$38} & 26{\tiny$\pm$22} & 7.3{\tiny$\pm$8.0} & $-$0.27{\tiny$\pm$0.70} \\
TCN & 818{\tiny$\pm$380} & 544{\tiny$\pm$252} & 36.6{\tiny$\pm$14.9} & 0.57{\tiny$\pm$0.18} & 26{\tiny$\pm$14} & 19{\tiny$\pm$10} & 23.7{\tiny$\pm$9.5} & 0.46{\tiny$\pm$0.29} & 38{\tiny$\pm$36} & 24{\tiny$\pm$19} & 6.9{\tiny$\pm$6.7} & $-$0.24{\tiny$\pm$0.66} \\
Transformer & 823{\tiny$\pm$331} & 548{\tiny$\pm$221} & 37.2{\tiny$\pm$15.9} & 0.55{\tiny$\pm$0.24} & 26{\tiny$\pm$15} & 19{\tiny$\pm$11} & 24.1{\tiny$\pm$9.2} & 0.45{\tiny$\pm$0.29} & 39{\tiny$\pm$36} & 26{\tiny$\pm$20} & 7.3{\tiny$\pm$7.4} & $-$0.30{\tiny$\pm$0.76} \\
Mamba & 916{\tiny$\pm$429} & 621{\tiny$\pm$299} & 40.2{\tiny$\pm$13.7} & 0.50{\tiny$\pm$0.26} & 31{\tiny$\pm$16} & 23{\tiny$\pm$13} & 28.0{\tiny$\pm$10.4} & 0.33{\tiny$\pm$0.41} & 44{\tiny$\pm$39} & 31{\tiny$\pm$25} & 8.9{\tiny$\pm$9.4} & $-$0.63{\tiny$\pm$1.78} \\
LSTM & 880{\tiny$\pm$399} & 597{\tiny$\pm$280} & 40.0{\tiny$\pm$14.8} & 0.52{\tiny$\pm$0.24} & 29{\tiny$\pm$14} & 22{\tiny$\pm$11} & 28.3{\tiny$\pm$10.4} & 0.36{\tiny$\pm$0.35} & 47{\tiny$\pm$41} & 32{\tiny$\pm$26} & 9.2{\tiny$\pm$9.7} & $-$0.73{\tiny$\pm$1.65} \\
\midrule
\multicolumn{13}{l}{\textit{Statistical baselines}} \\
NaiveLast & 1900{\tiny$\pm$812} & 1423{\tiny$\pm$611} & 84.6{\tiny$\pm$16.1} & 0.00{\tiny$\pm$0.00} & 51{\tiny$\pm$21} & 38{\tiny$\pm$16} & 47.7{\tiny$\pm$16.5} & 0.00{\tiny$\pm$0.00} & 35{\tiny$\pm$36} & \textbf{19}{\tiny$\pm$16} & \textbf{5.4}{\tiny$\pm$5.4} & 0.00{\tiny$\pm$0.00} \\
MovingAvg & 1359{\tiny$\pm$592} & 1123{\tiny$\pm$446} & 71.1{\tiny$\pm$15.7} & 0.28{\tiny$\pm$0.02} & 38{\tiny$\pm$16} & 31{\tiny$\pm$13} & 38.8{\tiny$\pm$13.9} & 0.25{\tiny$\pm$0.07} & 38{\tiny$\pm$38} & 23{\tiny$\pm$20} & 6.5{\tiny$\pm$6.7} & $-$0.08{\tiny$\pm$0.07} \\
ARIMA & 1543{\tiny$\pm$685} & 1210{\tiny$\pm$479} & 76.9{\tiny$\pm$15.9} & 0.19{\tiny$\pm$0.20} & 40{\tiny$\pm$19} & 32{\tiny$\pm$14} & 40.2{\tiny$\pm$14.2} & 0.20{\tiny$\pm$0.10} & 35{\tiny$\pm$37} & 20{\tiny$\pm$17} & 5.5{\tiny$\pm$5.5} & 0.00{\tiny$\pm$0.02} \\
\bottomrule
\end{tabular*}
\end{table*}

\begin{table*}[t]
\centering
\caption{Personalised results on GLOBEM at $h$=8. Each model is fine-tuned per participant. TimesFM (zero-shot) included as reference.}
\label{tab:per_feature_h8_personal}
\scriptsize
\begin{tabular*}{\textwidth}{@{\extracolsep{\fill}} l rrrr rrrr rrrr}
\toprule
& \multicolumn{4}{c}{\textbf{Steps (daily count)}} & \multicolumn{4}{c}{\textbf{Screen (min/day)}} & \multicolumn{4}{c}{\textbf{Sleep (min/day)}} \\
\cmidrule(lr){2-5} \cmidrule(lr){6-9} \cmidrule(lr){10-13}
\textbf{Model} & RMSE & MAE & sMAPE & Skill & RMSE & MAE & sMAPE & Skill & RMSE & MAE & sMAPE & Skill \\
\midrule
\textit{TimesFM}$^\dagger$ & 1363{\tiny$\pm$580} & --- & --- & 0.28{\tiny$\pm$0.46} & 39{\tiny$\pm$19} & --- & --- & 0.23{\tiny$\pm$0.38} & 42{\tiny$\pm$53} & --- & --- & $-$0.20{\tiny$\pm$1.51} \\
\midrule
\multicolumn{13}{l}{\textit{Personalised (fine-tuned)}} \\
PatchTST & \textbf{512}{\tiny$\pm$130} & \textbf{352}{\tiny$\pm$141} & \textbf{19.3}{\tiny$\pm$7.8} & \textbf{0.73}{\tiny$\pm$0.12} & \textbf{19}{\tiny$\pm$15} & \textbf{14}{\tiny$\pm$12} & \textbf{16.9}{\tiny$\pm$5.4} & \textbf{0.62}{\tiny$\pm$0.31} & 22{\tiny$\pm$15} & \textbf{15}{\tiny$\pm$9} & \textbf{3.9}{\tiny$\pm$2.7} & 0.39{\tiny$\pm$0.42} \\
MLP & 693{\tiny$\pm$191} & 487{\tiny$\pm$193} & 25.0{\tiny$\pm$8.3} & 0.64{\tiny$\pm$0.15} & 22{\tiny$\pm$18} & 17{\tiny$\pm$14} & 19.4{\tiny$\pm$6.0} & 0.57{\tiny$\pm$0.35} & \textbf{21}{\tiny$\pm$12} & 16{\tiny$\pm$9} & 4.2{\tiny$\pm$2.6} & \textbf{0.41}{\tiny$\pm$0.35} \\
TCN & 751{\tiny$\pm$269} & 532{\tiny$\pm$162} & 26.8{\tiny$\pm$10.3} & 0.60{\tiny$\pm$0.19} & 21{\tiny$\pm$15} & 16{\tiny$\pm$11} & 19.4{\tiny$\pm$6.7} & 0.59{\tiny$\pm$0.29} & 22{\tiny$\pm$12} & 16{\tiny$\pm$8} & 4.3{\tiny$\pm$2.3} & 0.37{\tiny$\pm$0.34} \\
Transformer & 735{\tiny$\pm$263} & 524{\tiny$\pm$153} & 26.9{\tiny$\pm$11.5} & 0.61{\tiny$\pm$0.19} & 21{\tiny$\pm$17} & 16{\tiny$\pm$13} & 20.3{\tiny$\pm$6.6} & 0.58{\tiny$\pm$0.34} & 21{\tiny$\pm$11} & 16{\tiny$\pm$8} & 4.2{\tiny$\pm$2.6} & 0.40{\tiny$\pm$0.33} \\
Mamba & 916{\tiny$\pm$519} & 648{\tiny$\pm$380} & 28.0{\tiny$\pm$14.1} & 0.52{\tiny$\pm$0.38} & 26{\tiny$\pm$22} & 20{\tiny$\pm$17} & 22.4{\tiny$\pm$14.0} & 0.49{\tiny$\pm$0.44} & 28{\tiny$\pm$18} & 22{\tiny$\pm$15} & 6.0{\tiny$\pm$4.5} & 0.20{\tiny$\pm$0.51} \\
LSTM & 943{\tiny$\pm$417} & 689{\tiny$\pm$285} & 32.8{\tiny$\pm$13.6} & 0.50{\tiny$\pm$0.27} & 28{\tiny$\pm$26} & 22{\tiny$\pm$21} & 24.3{\tiny$\pm$8.2} & 0.45{\tiny$\pm$0.51} & 23{\tiny$\pm$14} & 17{\tiny$\pm$11} & 4.6{\tiny$\pm$3.2} & 0.35{\tiny$\pm$0.40} \\
\bottomrule
\multicolumn{13}{l}{\footnotesize $^\dagger$Zero-shot foundation model (reference, not fine-tuned).}
\end{tabular*}
\end{table*}

Tables \ref{tab:per_feature_h8_global} present per-feature results on GLOBEM at short ($h$=1) and long ($h$=8) horizons. The three features showed different prediction difficulty. Step counts are hardest (best trained RMSE 673 at $h$=1), sleep duration is easiest (NaiveLast achieves 25 min RMSE due to sleep regularity) and screen time falls in between (best 18 min RMSE). PatchTST achieves the lowest RMSE among trained models for steps and screen at both horizons. TimesFM excels on sleep (10 min at $h$=1 vs.\ 26 for PatchTST) and matches trained models on steps, confirming its zero-shot strength. The skill score column reveals that all global trained models show negative skill on sleep, indicating that the feature's regularity makes NaiveLast a strong baseline, while personalised models achieve a positive skill (up to 0.59). This reflects the high correlation of individual sleep patterns, global models converge to population averages that overshoot or undershoot individual baselines, adding noise relative to the self-predictive NaiveLast. ARIMA, despite automatic parameter selection, performs only marginally above NaiveLast (skill 0.10-0.17 on steps), confirming the value of learned representations.

\begin{figure}[t]
\centering
\includegraphics[width=\linewidth]{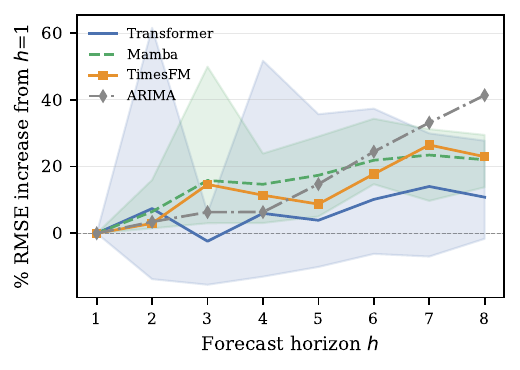}
\caption{Percentage RMSE increase relative to $h$=1 on GLOBEM. Trained models (Transformer, Mamba) show a mean$\pm$min-max band across context windows. TimesFM and ARIMA show mean across windows.}
\label{fig:rmse_degradation}
\end{figure}

Horizon degradation is feature-dependent: PatchTST degrades by $+$28\% RMSE from $h$=1 to $h$=8 on steps, while Mamba degrades by $+$25\% (Figure~\ref{fig:rmse_degradation}). At $h$=8, personalised PatchTST achieves the best RMSE on steps (512) andscreen (19~min), while personalised MLP leads on sleep (21~min) (Table~\ref{tab:per_feature_h8_personal}), showing that fine-tuning is most valuable at longer horizons where global models struggle.

\subsection{Personalisation}

Table~\ref{tab:per_feature_h8_personal} reports personalised results for all six trained architectures (200 evaluation participants, LR=$10^{-4}$, 200 epochs). The benefit varies substantially by feature: sleep shows the largest improvement (43-60\% RMSE reduction), followed by screen time (19-36\%) and step counts (16-46\%). PatchTST achieves the lowest personalised RMSE on steps (512 at $h$=8). Notably, personalisation ranking differs from global ranking: MLP and TCN, mid-tier globally, achieve competitive personalised performance (MLP sleep RMSE 21 vs.\ 39 global at $h$=8), while Transformer and LSTM show more modest gains.

\begin{figure}[t]
\centering
\includegraphics[width=\linewidth]{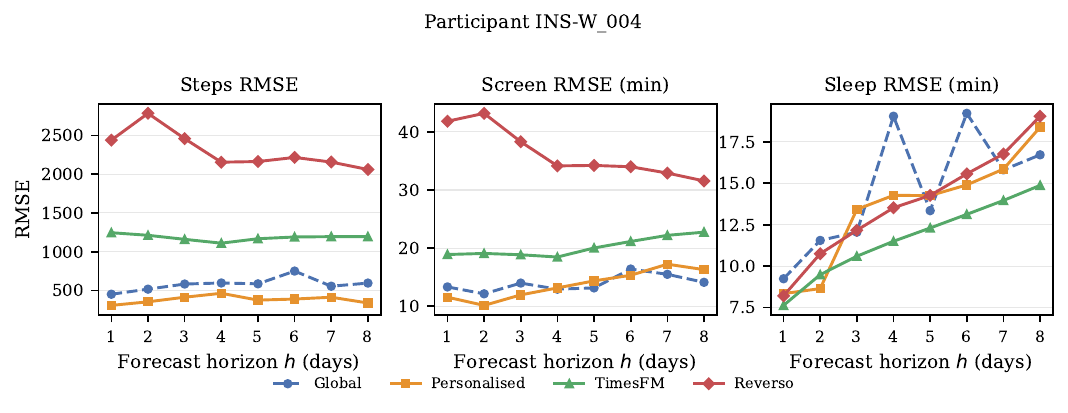}
\caption{Per-feature RMSE across horizons for participant INS-W\_004 (median RMSE). Four modelling approaches: Global Transformer, Personalised Transformer, TimesFM (zero-shot), and Reverso (zero-shot).}
\label{fig:per_person}
\end{figure}

Figure~\ref{fig:per_person} illustrates these trends for a representative participant (INS-W\_004, selected by median RMSE). Personalised predictions track the participant's behaviour more closely across all features and horizons, while the FMs (TimesFM, Reverso) show competitive short-horizon performance on sleep duration feature but degrade rapidly on steps.

Personalisation requires sufficient per-person data. On GLOBEM ($\sim$84 days per participant), fine-tuning is effective. On CAPTURE-24 ($\sim$22 hourly observations), personalisation helps only at $h$=1 but \emph{degrades} performance at longer horizons (up to $-$40\%), confirming that fine-tuning overfits when data is scarce.

\subsection{Foundation Models}
\label{sec:foundation_models}

To assess transferability, we evaluate all architectures across the three datasets. Tables \ref{tab:per_feature_h8_global} and \ref{tab:cross_dataset} reports skill scores relative to NaiveLast.

\begin{table}[t]
\centering
\caption{Cross-dataset sleep forecasting. CAPTURE-24: sleep (min/day),  $w$=8, $h$=8$^\dagger$, $N$=23. Wearables: sleep (hours/day), $w$=8, $h$=8, $N$=10. Trained models use $w$=8.}
\label{tab:sleep}
\small
\begin{tabular*}{\columnwidth}{@{\extracolsep{\fill}} l rr rr}
\toprule
& \multicolumn{2}{c}{\textbf{CAPTURE-24}} & \multicolumn{2}{c}{\textbf{Wearables}} \\
\cmidrule(lr){2-3} \cmidrule(lr){4-5}
\textbf{Model} & RMSE & Skill & RMSE & Skill \\
\midrule
\multicolumn{5}{l}{\textit{Foundation (zero-shot)}} \\
TimesFM & 30{\tiny$\pm$4} & $-$0.64{\tiny$\pm$0.58} & 14 & 0.15 \\
\midrule
\multicolumn{5}{l}{\textit{Global (trained)}} \\
PatchTST & 19{\tiny$\pm$6} & 0.04{\tiny$\pm$0.04} & --- & --- \\
MLP & --- & --- & 14{\tiny$\pm$0.50} & 1.3{\tiny$\pm$0.22} \\
TCN & --- & --- & \textbf{13}{\tiny$\pm$0.45} & \textbf{1.9}{\tiny$\pm$0.13} \\
Transformer & 20{\tiny$\pm$7} & $-$0.01{\tiny$\pm$0.04} & 15{\tiny$\pm$0.54} & 0.04{\tiny$\pm$0.27} \\
Mamba & \textbf{16}{\tiny$\pm$6} & \textbf{0.21}{\tiny$\pm$0.21} & 14{\tiny$\pm$0.49} & .14{\tiny$\pm$0.20} \\
LSTM & 20{\tiny$\pm$7} & 0.00{\tiny$\pm$0.04} & 13{\tiny$\pm$0.47} & 0.17{\tiny$\pm$0.19} \\
\midrule
\multicolumn{5}{l}{\textit{Statistical baselines}} \\
NaiveLast & 20{\tiny$\pm$7} & 0.00{\tiny$\pm$0.00} & 16{\tiny$\pm$0.52} & 0.00{\tiny$\pm$0.00} \\
\bottomrule
\end{tabular*}
\end{table}

\begin{table*}[t]
\centering
\caption{Cross-dataset comparison at maximum horizon. Values are mean{\tiny$\pm$}std of per-participant metrics, averaged across each dataset's features. GLOBEM: $h$=8, $D$=3, $N$=200. CAPTURE-24: $h$=6, $D$=4, $N$=23. Wearables: $h$=8, $D$=2, $N$=10. ``---'' indicates unavailability.$^\dagger$}
\label{tab:cross_dataset}
\scriptsize
\begin{tabular*}{\textwidth}{@{\extracolsep{\fill}} l rrrr rrrr rrrr}
\toprule
& \multicolumn{4}{c}{\textbf{GLOBEM ($h$=8)}} & \multicolumn{4}{c}{\textbf{CAPTURE-24 ($h$=6)}} & \multicolumn{4}{c}{\textbf{Wearables ($h$=8)}} \\
\cmidrule(lr){2-5} \cmidrule(lr){6-9} \cmidrule(lr){10-13}
\textbf{Model} & RMSE & MAE & sMAPE & Skill & RMSE & MAE & sMAPE & Skill & RMSE & MAE & sMAPE & Skill \\
\midrule
\multicolumn{13}{l}{\textit{Foundation (zero-shot)}} \\
TimesFM & 481{\tiny$\pm$117} & --- & --- & 0.11{\tiny$\pm$0.78} & 19{\tiny$\pm$5} & --- & --- & $-$0.07{\tiny$\pm$0.29} & 25 & 11 & 84 & 0.20 \\
Reverso & 719{\tiny$\pm$137} & --- & --- & $-$0.04{\tiny$\pm$0.77} & --- & --- & --- & --- & 42 & 16 & --- & $-$0.16 \\
\midrule
\multicolumn{13}{l}{\textit{Global (trained)}} \\
PatchTST & \textbf{229}{\tiny$\pm$97} & \textbf{140}{\tiny$\pm$120} & \textbf{17.2}{\tiny$\pm$9.5} & \textbf{0.42}{\tiny$\pm$0.15} & 15{\tiny$\pm$5} & 10{\tiny$\pm$3} & 151{\tiny$\pm$12} & 0.21{\tiny$\pm$0.09} & --- & --- & --- & --- \\
MLP & 300{\tiny$\pm$44} & 201{\tiny$\pm$161} & 22.8{\tiny$\pm$10.6} & 0.25{\tiny$\pm$0.39} & --- & --- & --- & --- & 27{\tiny$\pm$29} & 14{\tiny$\pm$15} & 88{\tiny$\pm$15} & 0.16{\tiny$\pm$0.19} \\
TCN & 294{\tiny$\pm$143} & 196{\tiny$\pm$160} & 22.4{\tiny$\pm$10.4} & 0.26{\tiny$\pm$0.38} & --- & --- & --- & --- & 27{\tiny$\pm$29} & 15{\tiny$\pm$16} & 87{\tiny$\pm$16} & 0.19{\tiny$\pm$0.15} \\
Transformer & 296{\tiny$\pm$127} & 198{\tiny$\pm$151} & 22.9{\tiny$\pm$10.8} & 0.23{\tiny$\pm$0.43} & 16{\tiny$\pm$5} & 10{\tiny$\pm$4} & 156{\tiny$\pm$11} & 0.18{\tiny$\pm$0.12} & 26{\tiny$\pm$27} & 14{\tiny$\pm$14} & 86{\tiny$\pm$17} & 0.13{\tiny$\pm$0.21} \\
Mamba & 330{\tiny$\pm$162} & 225{\tiny$\pm$179} & 25.7{\tiny$\pm$11.2} & 0.07{\tiny$\pm$0.82} & \textbf{14}{\tiny$\pm$5} & \textbf{9}{\tiny$\pm$3} & \textbf{149}{\tiny$\pm$13} & \textbf{0.26}{\tiny$\pm$0.16} & 26{\tiny$\pm$28} & 14{\tiny$\pm$15} & \textbf{83}{\tiny$\pm$19} & 0.18{\tiny$\pm$0.18} \\
LSTM & 319{\tiny$\pm$152} & 217{\tiny$\pm$172} & 25.8{\tiny$\pm$11.7} & 0.05{\tiny$\pm$0.74} & 16{\tiny$\pm$5} & 10{\tiny$\pm$4} & 156{\tiny$\pm$11} & 0.17{\tiny$\pm$0.12} & 27{\tiny$\pm$28} & 14{\tiny$\pm$15} & 87{\tiny$\pm$13} & \textbf{0.19}{\tiny$\pm$0.17} \\
\midrule
\multicolumn{13}{l}{\textit{Statistical baselines}} \\
NaiveLast & 662{\tiny$\pm$290} & 493{\tiny$\pm$281} & 45.9{\tiny$\pm$12.7} & 0.00{\tiny$\pm$0.00} & 18{\tiny$\pm$6} & 12{\tiny$\pm$4} & 80{\tiny$\pm$24} & 0.00{\tiny$\pm$0.00} & 34{\tiny$\pm$39} & 20{\tiny$\pm$23} & 63{\tiny$\pm$15} & 0.00{\tiny$\pm$0.00} \\
MovingAvg & 478{\tiny$\pm$282} & 392{\tiny$\pm$226} & 38.8{\tiny$\pm$12.1} & 0.15{\tiny$\pm$0.05} & 18{\tiny$\pm$6} & 13{\tiny$\pm$5} & 102{\tiny$\pm$30} & 0.06{\tiny$\pm$0.14} & 29{\tiny$\pm$32} & 21{\tiny$\pm$25} & 71{\tiny$\pm$16} & 0.17{\tiny$\pm$0.08} \\
ARIMA & 540{\tiny$\pm$247} & 421{\tiny$\pm$236} & 40.9{\tiny$\pm$11.9} & 0.13{\tiny$\pm$0.11} & 20{\tiny$\pm$7} & 13{\tiny$\pm$5} & 113{\tiny$\pm$32} & $-$0.07{\tiny$\pm$0.19} & --- & --- & --- & --- \\
\bottomrule
\multicolumn{13}{l}{\footnotesize $^\dagger$Foundation model MAE/sMAPE unavailable; Wearables foundation models evaluated on pooled data (no per-participant std).}\\
\multicolumn{13}{l}{\footnotesize CAPTURE-24 sMAPE is inflated by zero-inflated features (sleep, MVPA); RMSE and Skill are more reliable for that dataset.}
\end{tabular*}
\end{table*}

Table~\ref{tab:cross_dataset} reveals dataset-dependent architecture rankings. On GLOBEM, PatchTST leads across all metrics (RMSE 229, skill 0.42); TimesFM is competitive (RMSE 481) despite zero-shot inference, while Reverso trails (RMSE 719). On CAPTURE-24, Mamba achieves the highest skill (0.26) and lowest RMSE (14), while MLP and TCN diverge. TimesFM shows negative skill ($-$0.07), suggesting its pre-training underrepresents sub-daily health signals. On the small Wearables dataset ($N$=10), TimesFM (RMSE 25, skill 0.20) matches the best trained models, confirming FM value in low-data regimes. PatchTST fails on Wearables where sequences are too short for patch tokenisation. 

\begin{figure}[t]
\centering
\includegraphics[width=\linewidth]{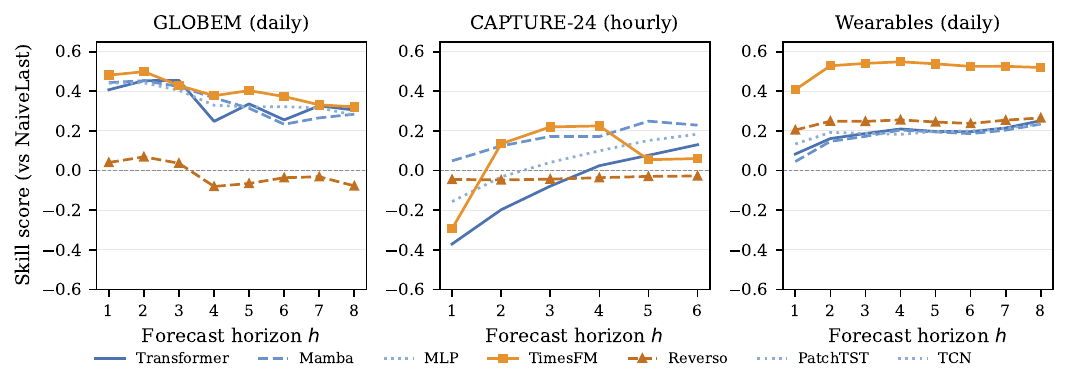}
\caption{Skill score (1 $-$ RMSE$_{\text{model}}$ / RMSE$_{\text{NaiveLast}}$) across forecast horizons on three datasets. Lines show skill at fixed context window $w$=8; FMs use their native context. The dashed grey line marks skill = 0 (NaiveLast baseline).}
\label{fig:skill_vs_horizon}
\end{figure}

Figure~\ref{fig:skill_vs_horizon} shows skill levels across horizons. On GLOBEM, trained models degrade from skill 0.40-0.50 at $h$=1 to 0.25-0.35 at $h$=8. On CAPTURE-24, the pattern reverses: most models show negative skill at $h$=1, worse than persistence, but recover to 0.15-0.25 by $h$=6 as NaiveLast's stale predictions become increasingly outdated; Mamba is the only model with positive skill throughout. On Wearables, TimesFM separates clearly from trained models (skill 0.40-0.55 vs.\ 0.05-0.25), confirming its advantage in low-data situations. Reverso remains near or below zero on all three datasets.

\subsection{Prediction Intervals}

\begin{figure}[t]
\centering
\includegraphics[width=\linewidth]{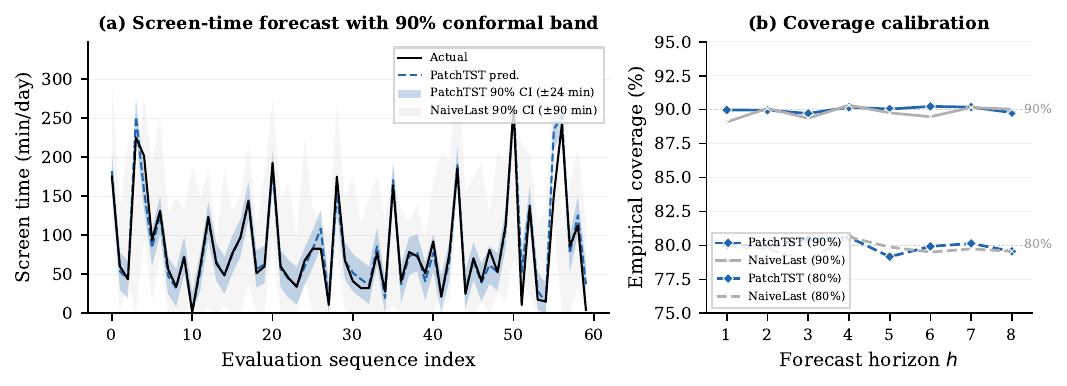}
\caption{Split prediction intervals on GLOBEM screen time ($w$=8). (a) One day ahead forecasts with 90\% conformal bands for PatchTST ($\pm$24 min) and NaiveLast ($\pm$90 min). (b) Empirical coverage at 80\% and 90\% nominal levels across horizons $h$=1-8. Both models achieve near-nominal coverage.}
\label{fig:intervals}
\end{figure}

To quantify forecast uncertainty, we construct empirical prediction intervals for screen time: evaluation sequences are randomly split 50/50 into calibration and test sets, absolute residuals are computed on the calibration set, and the $(1{-}\alpha)$-quantile defines symmetric prediction bands on the test set. Figure~\ref{fig:intervals}(a) shows that global PatchTST's 90\% prediction band is $\pm$24~min, 3.7$\times$ narrower than NaiveLast's $\pm$90~min, meaning a practitioner can tell a user ``your screen time tomorrow will be 93$\pm$24 minutes'' with 90\% confidence.

Personalised PatchTST achieves a comparable band ($\pm$25~min) on screen time, however, for step counts the personalised band is 9\% narrower at $h$=1 (1,483 vs.\ 1,625 steps), consistent with the larger RMSE improvements from personalisation on this feature. Figure~\ref{fig:intervals}(b) verifies that empirical coverage remains within 2\% of the 90\% nominal level across all horizons for both global and personalised models, confirming that the intervals are well-calibrated.

\subsection{Feature Importance}

\begin{table}[t]
\centering
\caption{Feature-group permutation importance on GLOBEM (Transformer, $w$=8, $h$=1, $N$=200, 5 repeats). Group importance is the RMSE increase when all features in a group are simultaneously permuted.$^\dagger$}
\label{tab:feature_importance}
\small
\begin{tabular*}{\columnwidth}{@{\extracolsep{\fill}} l r}
\toprule
\textbf{Feature group} & \textbf{Importance} \\
\midrule
steps & \textbf{243.0} \\
cross\_feature & 93.3 \\
screen\_min & 9.9 \\
sleep\_min & 7.7 \\
temporal & $-$0.2 \\
\bottomrule
\end{tabular*}
\end{table}

Table~\ref{tab:feature_importance} reports group-level permutation importance for a Transformer on GLOBEM at $h$=1. Step-count features contribute the highest group importance (243.0 RMSE delta) despite individually modest values, indicating a distributed predictive signal across participants. Cross-feature ( all features included sleep/screen/steps) rank third, confirming that engineered interactions carry meaningful information. Temporal encodings and metadata contribute near zero, suggesting that the model captures seasonality through the behavioural features themselves.

\section{Discussion}
\label{sec:discussion}

\textbf{Summary of findings.} Our results delineate three complementary modelling paradigms. Global trained models close 33-44\% of the NaiveLast gap on GLOBEM (Mamba skill 0.44 at $h$=1), with PatchTST leading on steps and screen time. The three runners-up (TCN, MLP, Transformer) show no practically meaningful differences in accuracy, suggesting that architecture choice matters less than personalisation strategy. TimesFM matches or exceeds trained architectures zero-shot (skill 0.48 at $h$=1) and dominates in low-data regimes (Wearables skill 0.41-0.55 vs.\ 0.05-0.28 for trained models), though its advantage depends on temporal granularity, with negative skill at $h$=1 on hourly CAPTURE-24 data. Personalised fine-tuning of all six architectures reduces per-feature RMSE by 16-60\%, with the benefit strongly dependent on feature regularity with sleep feature benefits most (43-60\%) due to stable individual patterns, while step counts benefit least (16-46\%).

\textbf{Architecture stability across datasets.}  Model rankings are not consistent across datasets, suggesting that inductive biases interact with data characteristics. PatchTST leads on GLOBEM (daily, $N$=200) where long sequences support effective patch tokenisation, but fails on Wearables where sequences are too short. Mamba is the only architecture with positive skill on hourly CAPTURE-24 data (0.11 at $h$=1), likely because its linear-complexity state-space formulation handles the higher temporal resolution without overfitting. On the small Wearables dataset ($N$=10), TCN achieves the best trained-model skill, consistent with its lower parameter count (98K vs.\ 1.7-2.0M for Transformer and Mamba). These shifts indicate that no single architecture can be recommended unconditionally: dataset size, temporal granularity and sequence length should guide model selection.

\textbf{Personalisation is horizon-dependent.} The benefit of personalisation grows with forecast horizon. At $h$=1, fine-tuning PatchTST yields only marginal improvements (3\% RMSE reduction on steps, 11\% on screen), because global models already capture short-term population-level patterns well, leaving little room for per-user gains. At $h$=8, the same model reduces RMSE by 28\% on steps and 30\% on screen, a large and consistent effect across participants. This pattern suggests that personalisation is most valuable precisely at longer horizons where global models struggle to capture individual behavioural drift. Importantly, personalisation can also hurt: LSTM sleep RMSE worsens by 31\% at $h$=1, a clear sign of overfitting when per-user data are limited and the global model already fits the feature well.

\textbf{Foundation model domain fit and computational cost.} The contrast between TimesFM and Reverso (skill 0.28–0.48 vs.\ $-$0.09–0.12 on GLOBEM) shows zero-shot transfer is highly sensitive to training data. TimesFM, pre-trained on over 100 billion parameters, generalises well to daily health features, whereas Reverso is less suited to the irregular, zero-inflated sleep duration distributions. Yet TimesFM performs poorly on hourly CAPTURE-24 data ($-$0.03 at $h$=1), indicating its pre-training underrepresents sub-daily health signals. Thus, zero-shot performance depends on pre-training–target domain alignment, not model size alone. Cost profiles also differ: our trained architectures (98K–2.0M parameters; Table~\ref{tab:complexity}) run inference in under one second on a modern GPU, while TimesFM (200M parameters) needs about 1\,GB of GPU memory at half precision, making on-device wearable inference impractical. Architecture choice should therefore weigh domain fit and compute budget alongside accuracy.

\textbf{Practical considerations.} Personalised sleep predictions reach 11–17 min RMSE and screen-time 14–19 min at h=1, suitable for threshold-based alerting. However, personalisation requires at least 1-2 weeks of individual data: on GLOBEM ($\sim$84 days per participant) fine-tuning is effective, while on CAPTURE-24 ($\sim$22 hourly observations) it degrades performance by up to 40\% at longer horizons. For practitioners, we suggest a decision framework starting with using FM (TimesFM) as a strong zero-shot default when per-user data are scarce or absent. Second, adopt PatchTST or Transformer for longitudinal datasets with sufficient training data and lastly apply personalised fine-tuning selectively at longer horizons where global models already show degraded skill.

\section{Conclusion}
\label{sec:conclusion}

This paper benchmarked nine forecasting approaches across three wearable datasets encompassing over 800 participants. Three findings emerge: (i) no single architecture dominates, though PatchTST leads among trained models, (ii) participant-level fine-tuning reduces per-feature RMSE by 16-60\%, with gains largest for sleep and at longer horizons and (iii) the FM TimesFM matches or exceeds trained models zero-shot, particularly in low-data regimes.

Our study has limitations: personalisation was evaluated on a 200-participant GLOBEM subset, FM evaluation was limited to zero-shot mode, and all datasets originate from relatively homogeneous populations. Future directions include federated personalisation~\cite{mcmahan2017fedavg, chen2020fedhealth}, fine-tuning FMs on health-specific corpora~\cite{das2024timesfm, fu2026reverso}, and probabilistic forecasting with uncertainty quantification~\cite{salinas2020deepar}.

\bibliographystyle{unsrt}
\bibliography{ref.bib} 

\end{document}